\documentclass{article}
\usepackage{spconf,amsmath,graphicx}

\title{Learning by Inertia: Self-supervised  Monocular Visual Odometry for Road Vehicles}
\name{Chengze Wang, Yuan Yuan$^{*}$, Qi Wang\thanks{$^*$Corresponding author. 
This work was supported by the National Natural Science Foundation of China under Grant U1864204 and 61773316, State Key Program of National Natural Science Foundation of China under Grant 61632018, Natural Science Foundation of Shaanxi Province under Grant 2018KJXX-024, Projects of Special Zone for National Defense Science and Technology Innovation, Fundamental Research Funds for the Central Universities under Grant 3102017AX010, and Open Research Fund of Key Laboratory of Spectral Imaging Technology, Chinese Academy of Sciences.}}
\address{School of Computer Science and Center for OPTical IMagery Analysis and Learning (OPTIMAL),\\
	Northwestern Polytechnical University, Xi'an 710072, Shaanxi, P. R. China\\}
\begin{document}
%
\maketitle
\begin{abstract}
In this paper, we present iDVO (inertia-embedded deep visual odometry), a self-supervised learning based monocular visual odometry (VO) for road vehicles. When modelling the geometric consistency within adjacent frames, most deep VO methods ignore the temporal continuity of the camera pose, which results in a very severe jagged fluctuation in the velocity curves. With the observation that road vehicles tend to perform smooth dynamic characteristics in most of the time, we design the inertia loss function to describe the abnormal motion variation, which assists the model to learn the consecutiveness from long-term camera ego-motion. Based on the recurrent convolutional neural network (RCNN) architecture, our method implicitly models the dynamics of road vehicles and the temporal consecutiveness by the extended Long Short-Term Memory (LSTM) block. Furthermore, we develop the dynamic hard-edge mask to handle the non-consistency in fast camera motion by blocking the boundary part and which generates more efficiency in the whole non-consistency mask. The proposed method is evaluated on the KITTI dataset, and the results demonstrate state-of-the-art performance with respect to other monocular deep VO and SLAM approaches.

\end{abstract}
\section{Introduction}
\label{intro}

Simultaneous localization and mapping (SLAM) is one of the critical capabilities for robots and self-driving vehicles to navigate in no GPS or RTK (Real-Time Kinematic) available environment. VO is the front-end module of a typical visual SLAM, which uses visual information to preliminarily estimate each pixel's depth and camera motion in the adjacent frames.  

In the past decades, multiple selections of sensors have been explored in the VO area, such as monocular camera, stereo camera, RGB-Depth, LiDAR, and visual-IMU (inertia measure unit). The monocular VO is under most investigations for its ubiquity and low cost.

For the monocular VO, model-based (or geometric) VO have been widely researched \cite{mur2017orb,newcombe2011dtam,engel2018direct}. However, some disadvantages of these methods are insurmountable, e.g., it is hard to handle non-consistency (non-rigid) structure and not robust in the high dynamic situation.

\begin{figure}[t]
\centering 
\includegraphics[width=8.5cm]{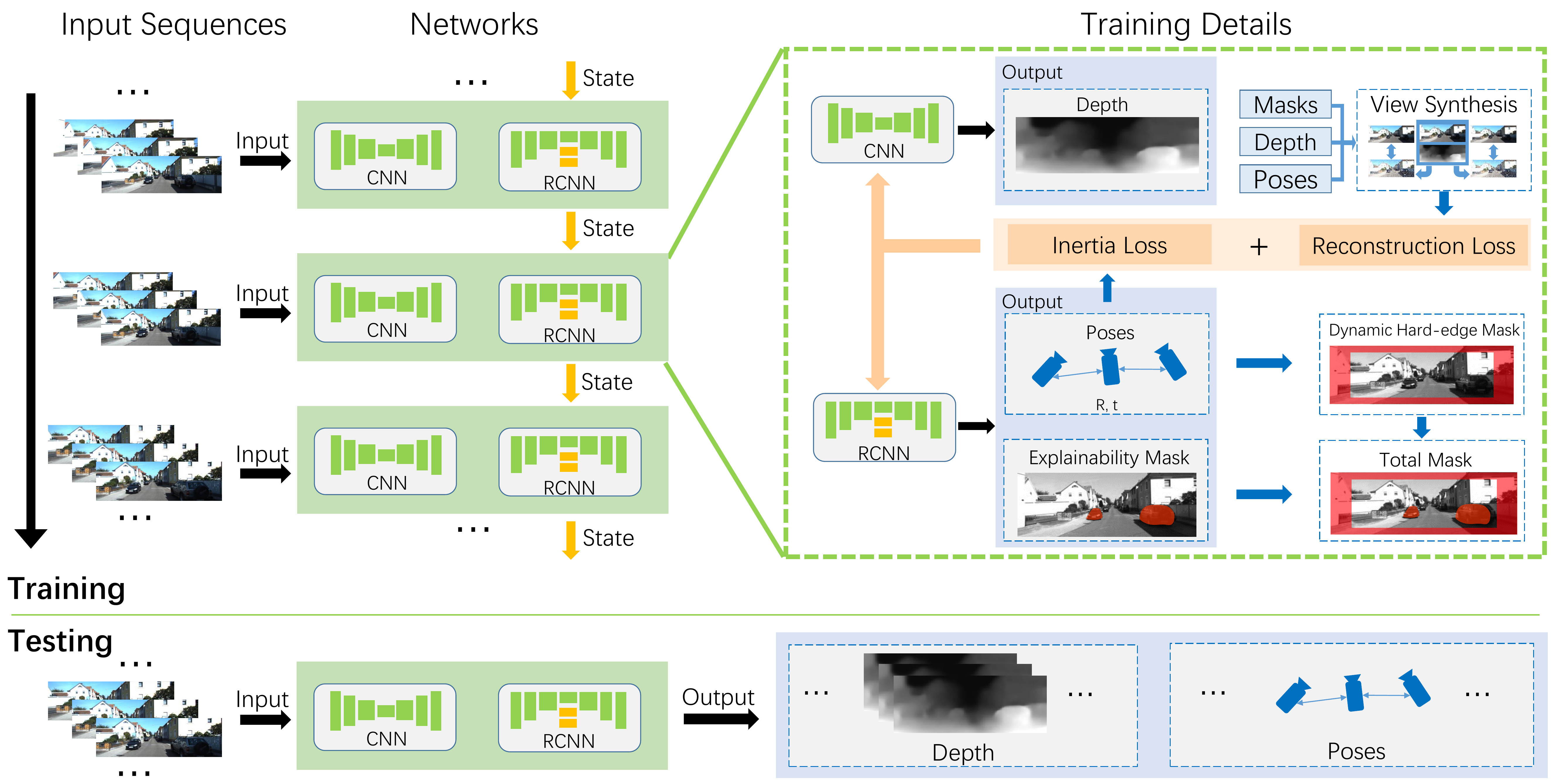} 
\label{fig1}
\caption{The overview of iDVO. The CNN-RCNN dual networks structure takes the sequences as input, and outputs the per-pixel depth with 6DoF poses sequentially. The total mask used in view synthesis is combined by the computed dynamic hard-edge mask and the estimated explain-ability mask. Both networks can be tested individually.}
\end{figure}
To overcome the limitations as mentioned above, recent deep learning based VO has been implemented and achieved a considerable performance against traditional methods. 
One key shortcoming of deep VO is that the collection of the ground truth is expensive and laborious \cite{geiger2013vision}, so the self-supervised methods which mainly based on the view synthesis \cite{szeliski1999prediction} emerge. 

However, most existing deep VO are imposing constraints on the feature/pixel-level error between adjacent frames \cite{eigen2014depth,ummenhofer2017demon,godard2017unsupervised}, or by the view synthesis \cite{zhou2017unsupervised,vijayanarasimhan2017sfm}. 
These VO estimate ego-motion only by the spatial information existing in several frames, which means temporal information within the frames is not fully utilized. As a result, the output of deep VO is inaccurate and discontinuous. One of the most obvious results is that the estimated speed curves (calculate by poses) present a severe saw-tooth undulation. 

Actually, with the effect of inertia, the motion variation of road vehicles is usually smooth, which means acceleration value is limited in a certain range. 
Under the normal driving situation, the vehicle's acceleration value is usually under 0.3g (gravitational acceleration) \cite{kuderer2015learning}, and the deceleration caused by nonurgent braking is mostly less than 0.2g \cite{eboli2016combining,mehar2013speed}. In the following part of this paper, we will use the word ``inertia" to describe these attributes of the motion status of vehicles. 

Focusing on the dynamic characteristic caused by inertia, we proposed the iDVO, a self-learning based inertia-embedded deep visual odometry, as illustrated in Fig. 1. 

The main contributions of this work are the following:

1) We propose the inertia loss to introduce the abnormal variation of motion status. This new constraint enables the model to learn from temporal and spatial consecutiveness within the vehicle motion simultaneously.

2) We extend the full convolutional architecture to RCNN while maintaining the system self-supervised. This RCNN is capable to automatically learn sequences' internal coupling.

3) Dynamic hard-edge mask is proposed to handle the massive viewpoint displacement in high-speed motion. 

To the best of our knowledge, this is the first approach that imposes the dynamic characteristic for the deep VO. It is possible to train with any forward-looking driving video. The experimental results on KITTI \cite{Geiger2012CVPR} Odometry datasets have verified the effectiveness of iDVO.

\section{Our Method}
\label{sec:Method}

Our method uses deep neural networks to learn the sequential ego-motion and the depth of frames from unlabelled dynamic driving video. 
On the network architecture, we introduce RNN block to our models to enhance the ability in dealing with temporal information. To treat with the non-consistency or dynamic structure in the frames, dynamic hard-edge mask is proposed to block unnecessary part of the image in projection.
Lastly, the inertial attribution function is introduced, and the relevant spatial constraints are also discussed.

\subsection{Network Architecture}
\label{net}

The encoder-decoder architecture based on DispNet \cite{mayer2016large} has been proven to work efficiently in deep VO.  This full-convolutional architecture automatically learns the features for ego-motion and depth estimation. However, as we mentioned in Sec. \ref{intro}, CNN focuses on the limited input frames, and ignores the temporal continuity in the whole shot. Therefore, in our approach, we adopt RCNN architecture to implement an end-to-end deep VO equipped with the ability to keep the continuity in ego-motion.

\begin{figure}[!htb]
\centering 
\includegraphics[width=8.5cm]{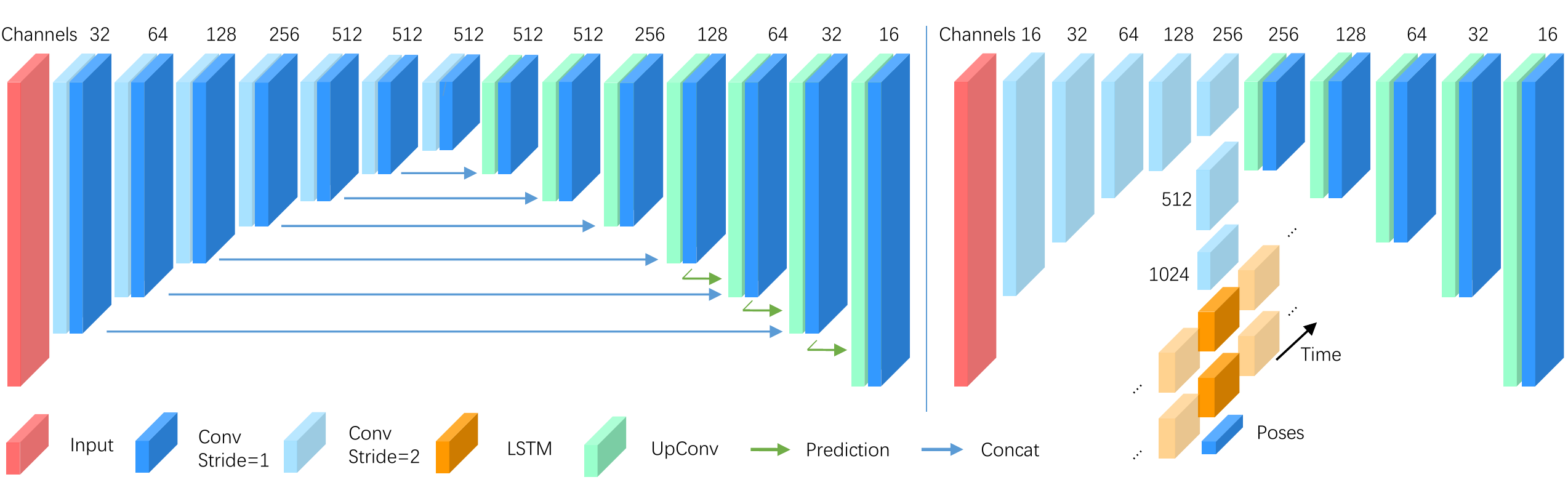} 
\caption{The CNN-RCNN network structure. The correspondence between color and layer/operation is shown in the legend at the bottom. {\it The left one}: the DispNet \cite{mayer2016large} architecture is adopted for the depth estimator CNN. {\it The right one}: The pose-prediction RCNN. The decoder part is for multi-scale explain-ability mask prediction.}
 \label{Networks}
\end{figure}

Inspired by SfMLearner \cite{zhou2017unsupervised}, our networks have two independent networks in form of encoder-decoder as shown in Fig. \ref{Networks}, one for the depth inferring, and the other one for the pose estimating and explain-ability mask generation.  
The depth inferring network is fully convolutional, estimating each pixel's depth of the input single image. For the first 4 convolutional layers, the kernel size is set to 7, 7, 5, 5, while all the other convolutional layers use the size of 3.   

To learn the connections in the sequence of video, we adopt the RCNN structure to the pose estimator network. Due to the regular RNN is hard to train on long time sequence, we choose to align the LSTM units with the original full convolutional network. Specifically, The ego-motion estimating and explain-ability mask generating network is extended with 2 layers of LSTM cells at the end of ego-motion estimating procedure. 
The LSTM cells input with the feature map generated by the fifth convolutional layer, and output pairs of relative poses in 6-DoF, and each of the LSTM layers has 1000 hidden states.
The multi-scale explain-ability mask is generated by the last 4 convolutional layers in the decoder part.
During the training, to maintain the order of sequence frames in every single shot, we abandon the file-level shuffle and replace it with a shot-level random selection.

\subsection{Dynamic Hard-edge Mask}
\label{DHEM}

Because our networks are trained by the video captured in dynamic vehicles (we remove the static frames by optical flow from the sequences), the view point will differ time by time,  which means some obstacles at the edges in one frame might not be captured in the next frame (when forward moving). As a result, in the reconstruction procedure, some pixels at the edge part of the resource frame will not be mapped in the boundary of the target frame. 

\begin{figure}[htb]
\centering
\includegraphics[width=8.6cm]{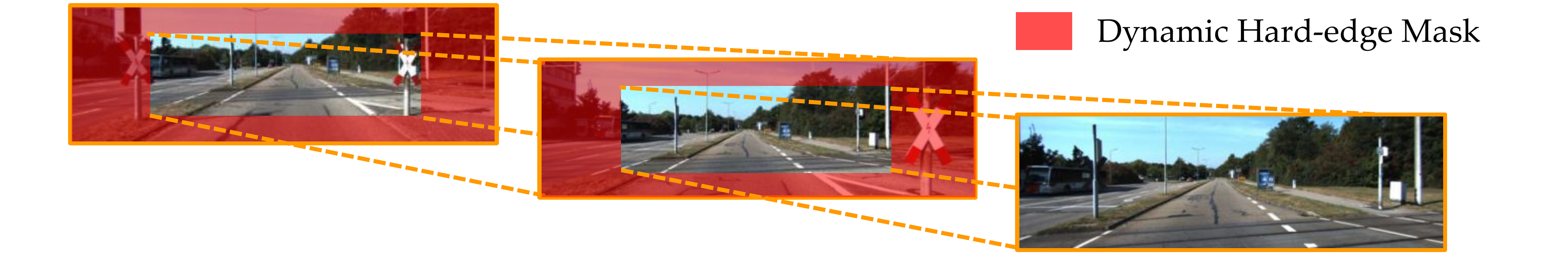} 
\caption{The sketch of dynamic hard-edge mask (the red part). For better display, the frame-to-frame time between these 3 frames is $\sim$0.5s, so the mask is much larger than actual mask we used in experiments.}
\label{mask}
\end{figure}

To generate the non-consistency mask efficiently and precisely, we propose the dynamic hard-edge mask (DHEM). The DHEM $M_t^{\rm{h}}$ is a hard mask in the edge of resource frame (as shown in Fig. \ref{mask}), which together with the soft estimated explain-ability mask $M_t^{\rm{e}}$ will form a complete mask ${M_t}$ at moment $t$. 
The hard mask blocks the edge pixels by ignoring the pixels in the mask during reconstruction. The edge mask's width of the mask is a positive correlation to the velocity at the frame-taken moment, which is calculated by the estimated pose from the ego-motion RCNN. 
The width of the left and right borders will be moderately adjusted according to the instant steering amplitude.

This efficient DHEM completes non-consistency mask with the explain-ability mask for the accurate view synthesis, which helps the training converge, also enhances the performance of depth and ego-motion estimation.

\subsection{Loss Functions}
\subsubsection{Inertia Loss Construction}

The inertia loss is constructed by the estimated poses in a period of time, more precisely it is calculated by the acceleration and jerk (the deviation of acceleration). 
We assume the vehicle accelerate and decelerate smoothly in the video capture procedure, and jerk of them also varies gently in a reasonable range. The input of our VO is sequential video captured on riding road vehicle. 
At the moment ${t}$ , the captured frame is ${F_t}$, and the depth network estimates its corresponding depth map ${D_t}$. Then the ego-motion estimation network estimates the relative 6-DoF poses ${T_{t}}$ and ${T_{t+1}}$ between adjacent frames ${F_{t-1}}$, ${F_t}$ and ${F_{t+1}}$ (if using three-frame snippet). Every relative pose ${T}$ is constructed by position ${P}$ and orientation ${\varphi}$.

The inertia loss is calculated by the estimated poses ${T_{(t_0, ... , t, t+1,t+2, ...)}}$ in one single shot. Here we represent the displacement as velocity ${v_t}$ for approximate fixed frame-to-frame time ($\sim$0.1s in KITTI). In loss calculating,  ${\varphi}$ is in the form of Euler angles to avoid the possible optimization problem in learning. The straight-line velocity and angular velocity are calculated individually by:
\begin{equation}
\begin{array}{l}
{v_t} = \left\| {{P_t} - {P_{t - 1}}} \right\| , {\varphi_t} = \left\| {{\varphi _t} - {\varphi _{t - 1}}} \right\| .
\end{array}
\label{eq:1}
\end{equation}
Then, the sum of both velocity difference form the ``acceleration" value ${a_t}$ as follows:
\begin{equation}
\begin{array}{l}
{a_t} = ({v_t} - {v_{t - 1}}) + \chi ({\varphi _t} - {\varphi _{t - 1}}) ,
\end{array}\
\label{eq:2}
\end{equation}
where $\chi$ is a scale factor to balance the angular acceleration.  The ``acceleration" value here is a scalar to describe the variation of the vehicle motion, rather than the exact acceleration vector. And the jerk ${j_t}$ is simply calculated by:
${j_t} = {a_t} - {a_{t - 1}}$.
Because both ${a_t}$ and ${j_t}$ have a reasonable range, we ignore the value in loss if their values less than their typical value $a_{typ}$  and  $j_{typ}$, respectively:
\begin{equation}
\begin{array}{l}
{L_{acce}}({T_t}) = \max (0,\frac{{\left| {{a_t}} \right| - {a_{typ}}}}{{{a_t}}})\\
{L_{jerk}}({T_t}) = \max (0,\frac{{\left| {{j_t}} \right| - {j_{typ}}}}{{{j_t}}}) .
\end{array}
\label{eq:4}
\end{equation}
Finally, the entire inertia loss becomes:
\begin{equation}
{L_{i}} = \sum\limits_t {\left| {{L_{acce}}({T_t}) + {L_{jerk}}({T_t})} \right|} .
\label{eq:5}
\end{equation}

\subsubsection{Spatial Losses Construction}
The main supervision in our method is still the view synthesis: given a frame shot in a scene, synthesize another frame in a different pose. In our learning procedure, given a pair of images ${F_t}$ and ${F_{t-1}}$, we use the networks to estimate the ego-motion ${T_t}$ and the depth ${D_t}$ and ${D_{t-1}}$. The depth ${D_t}$ can be projected as a point cloud ${C_t}$ in the view position of ${F_t}$, so does the ${D_{t-1}}$. 
By using the spatial transformer \cite{jaderberg2015spatial}, we can synthesize the reconstructed frames ${F'_t}$ and ${F'_{t-1}}$. Comparing the reconstructed frames with the target frames, we can propose the differentiable image reconstruction loss ${L_{rec}}$:

\begin{equation}
{L_{rec}} = \sum\limits_{u,v} {\left\| {(F_t^{uv} - F_t^{'uv})M_t^{uv}} \right\|}
\label{eq:7}
\end{equation}
where ${p_t}({u},{v})$ is one pixel in ${F_{t}}$ ,and the ${M_t}$ is the non-consistency mask as we mentioned in Sec. \ref{DHEM}. To prevent the non-consistency mask ${M_t}$ from minimizing to 0, the mask loss $L_{mask}$ is implemented in our work, which is the binary cross-entropy loss between the mask and a same-size frame with constant 1 value at every pixel. 

In order to comprehensively compare the reconstructed frame ${F'_t}$ with the target frame ${F_{t}}$,  structured similarity (SSIM) loss ${L_{SSIM}}$ \cite{wang2004image} and 3D point cloud alignment loss ${L_{3D}}$ \cite{mahjourian2018unsupervised} are also introduced as additional loss. 
\\
\textbf{Final Loss} All loss functions in this section are computed at 4 different scales $s$, which are $\frac{1}{8}$, $\frac{1}{4}$, $\frac{1}{2}$, and original input resolution. The total loss is:
\begin{equation}
\footnotesize
{L_{total}} = \omega_1 {L_{i}} + \sum\limits_s {(\omega_2 {L_{rec}^s} + \omega_3 {L_{SSIM}^s} + \omega_4 {L_{3D}^s}+ \omega_5 {L_{mask}^s})} .
\label{eq:10}
\end{equation}
The total loss leads the model to learn the VO task in 3 perspectives. 
The inertia loss demonstrates the temporal constraint, while the SSIM loss and 3D loss demonstrate the spatial constraint. The view synthesis loss and the mask loss incarnate temporal and spatial constraint simultaneously.

\section{Experiments}
\label{sec:Exp}

In this section, we evaluate the proposed iDVO system with several state-of-the-art VO and SLAM systems in two perspective, depth estimation and pose estimation.
\subsection{Datasets and Implementation Details}

The KITTI dataset \cite{Geiger2012CVPR} is the most common benchmark for VO algorithm in the transportation scene. 
During training the frames are cropped and resized to 416 ${\times}$ 128, and multiple images are stacked to one snippet as ego-motion RCNN's input. Data augmentation including rescaling, cropping and luminance correction is applied to enlarge the KITTI dataset.
In training, we remove static frames automatically by limiting the optical flow between 2 frames.

We use Pytorch framework to implement our networks, and the optimizer is set to Adam with ${\beta_1}$ = 0.9, ${\beta_2}$ = 0.999, learning rate ${\alpha_2}$ = 0.0002. Typical values $a_{typ}$ and $j_{typ}$ equal to 2.0, 0.5, and batch-size is set to 16. And hyper-parameters $\chi$, $ \omega_1$, $\omega_2$, $\omega_3$, $\omega_4$ and $\omega_5$ are determinded empirically, which are respectively set to 100, 1, 1, 0.2, 0.1, 0.15. Due to using typical values $a_{typ}$ and $j_{typ}$, the depth estimator starts training with pre-trained model (10k iteration, SfMLearner) to make it easier to converge. 

\subsection{Pose Estimation Evaluation}

In order to facilitate evaluation results, we choose KITTI's monocular sequences 09 and 10 for pose estimation evaluation. 
ORB-SLAM \cite{mur2017orb} is a well-known SLAM system with loop closure and re-localization function, which indicates that the full version ORB-SLAM use the whole video sequences as supervision. ORB-SLAM (short) is supervised by 5 frame snippets, as same as our method's input.  
The baseline is the SfMLearner \cite{zhou2017unsupervised} trained with Cityscapes \cite{cordts2016cityscapes} and KITTI dual datasets. 
Futher more, we also compare our iDVO to the enhanced-SfMLearner proposed by Mahjourian et al. \cite{mahjourian2018unsupervised} , which represents the state-of-the-art performance.
Similar to \cite{zhou2017unsupervised}, we need to solve the scale factor by post-processing. Absolute Trajectory Error (ATE) for all methods is computed on 5-snippets and then averages over the entire sequence. The results of motion estimation are presented in Table \ref{pose1}. 
The results show that our method outperforms both the state-of-the-art deep VO approach and full SLAM system on monocular camera. 
More importantly, the smaller variance proves that our method has better stability on long-term ego-motion, which is critical in autonomous driving application.

To analyze the effectiveness of different components, we test three variants of our iDVO. 1) On non-consistency mask, we only use explain-ability mask. 2) The inertia loss is dropped. 3) The LSTM cells to predict poses in RCNN are replaced by the CNN blocks with kernel size of 1. The results of ablation study in Table \ref{pose2}. show all three modification are essential to enhance the performance. 

\begin{table}[htb!]
\footnotesize
\begin{center}
\caption{
Absolute Trajectory Error (ATE) results on KITTI VO Sequence 09 and 10 (lower is better). 
}
\label{pose1}
\begin{tabular}{|l|c|c|}		
			\hline
			 \textbf{Method} &  Seq. 09 & Seq. 10  \\
			\hline
			ORB-SLAM (full) &  0.014$\pm$0.008  & 0.012$\pm$0.011 \\
			\hline
			ORB-SLAM (short) & 0.064$\pm$0.141  & 0.061$\pm$0.130  \\
			Zhou et al. \cite{zhou2017unsupervised} &0.021$\pm$0.017 &0.020$\pm$0.015\\
			Mahjourian et al. \cite{mahjourian2018unsupervised} & 0.013$\pm$0.010 & 0.012$\pm$0.011\\
				
			\hline
		 Our iDVO & \textbf{0.012$\pm$0.011}& \textbf{0.012$\pm$0.010} \\
			
			\hline
		\end{tabular}
\end{center}
\vspace{-0.5cm}
\end{table}

\begin{table}[h]
\footnotesize
\begin{center}
\caption{
Absolute Trajectory Error (ATE) results of ablation experiments. 
}
\label{pose2}
\begin{tabular}{|l|c|c|}		
			\hline
			 \textbf{Method} &  Seq. 09 & Seq. 10  \\
	\hline
			Full iDVO & \textbf{0.012$\pm$0.011}& \textbf{0.012$\pm$0.010}  \\
			\hline
			w/o DHEM& 0.013$\pm$0.012  & 0.014$\pm$0.010  \\
			w/o Inertia loss & 0.016$\pm$0.014 &0.015$\pm$0.015\\
			w/o RCNN & 0.014$\pm$0.016 &0.013$\pm$0.014\\
			
			\hline
		\end{tabular}
\end{center}
\vspace{-0.5cm}
\end{table}

\subsection{Depth Estimation Evaluation}

The depth estimation evaluation which using the Eigen \cite{eigen2014depth} test set. Table \ref{tab111} quantitatively compares the depth estimation between iDVO and several state-of-the-art deep VO systems \cite{liu2016learning,zhou2017unsupervised,mahjourian2018unsupervised,li2017undeepvo}. 
Table \ref{tab111} shows that our method achieves the best performance except for Sq Rel section, and performs a big lead in RMSE section.
On the whole, our method has a great improvement than state-of-the-art algorithms in depth estimation, and also proves that constraints we have added to the ego-motion estimation can be effectively reflected in the depth estimation.

\begin{table}[h]
\begin{center}

\caption{
Single-view depth evaluation results on the KITTI Eigen et al. \cite{eigen2014depth} set.
}
\label{tab111}
\footnotesize
{
\begin{tabular}{|lcccc|}
			\hline
			 \textbf{Method}  &Abs Rel&Sq Rel& RMSE & RMSE log  \\
		
			\hline
			Eigen et al. \cite{eigen2014depth}             &0.203& 1.548& 6.307& 0.282\\
			Liu et al.  \cite{liu2016learning}        & 0.202 &1.614 &6.523& 0.275\\

			Zhou et al. \cite{zhou2017unsupervised} &0.208& 1.768& 6.856& 0.283 \\ 
			Mahjourian et al. \cite{mahjourian2018unsupervised}  &  0.163 &  \textbf{1.240} & 6.220 & 0.250  \\		
			Li et al. \cite{li2017undeepvo}     &    0.183 &1.73& 6.570& 0.268 		\\
			\hline
			\textbf{Our iDVO}                    &\textbf{ 0.162} & 1.277 & \textbf{5.114}  & \textbf{0.248}  \\			

			\hline
		\end{tabular}
		}
\end{center}
\vspace{-0.5cm}
\end{table}

\section{Conclusion}
\label{sec:conclu}

Most existing unsupervised deep VO optimize errors by using the image-based constraint such as the inaccuracy and coarseness of the reconstructed frame. These methods lack considerations of the temporal continuity of the camera pose, resulting in a severe saw-tooth undulation in the velocity curves.
Based on the prior of road vehicles' dynamic characteristics, we present a loss function to describe the abnormal estimated motion, and the results proved that it is effective to build constraints on long-term sequential ego-motion to supervise the training.
Also, the RCNN and the DHEM improve the performance of the algorithm from temporal and spatial perspective respectively. Experimental results show that our iDVO achieves state-of-the-art performance. 

\bibliographystyle{IEEEbib}
\bibliography{strings,refs}

\end{document}